%% file: main.tex
\documentclass[letterpaper, 10 pt, conference]{ieeeconf}

\usepackage{graphicx}
\graphicspath{ {./Figures/} }
\IEEEoverridecommandlockouts
\input{Preamble}
\begin{document}
\input{Acronym}

\title{Making Waves: A Membrane-Coupled Delta Array for Manipulating Objects Below the Actuator Spacing\\}

\author{
    Bailey Dacre, Andrés Faíña, Oliver Kroemer, Zeynep Temel \\
    Email:\mailto{ \{baid, anfv\}@itu.dk, \{okroemer, ztemel\}@andrew.cmu.edu}
}

\maketitle

\input{Sections/0_Abstract}


\input{Sections/1_Introduction}

\input{Sections/2_RelatedWork}
\input{Sections/3_MaterialsAndMethods}

\input{Sections/4_WorkspaceAnalysis}

\input{Sections/5_FieldsAndPrimatives}
\input{Sections/7_ReinfocmentLearning}
\input{Sections/8_New_Disscussion}

\input{Sections/9_Conclusion}

\bibliography{delta_references}
\bibliographystyle{IEEEtran}
\end{document}

%% file: preamble.tex
\usepackage[UKenglish]{babel}
\usepackage[utf8]{inputenc}
\usepackage[T1]{fontenc}
\usepackage{newfloat}
\usepackage{booktabs}
\usepackage{amssymb, amsmath, array, bm, algorithm, standalone,csquotes,textcomp, amssymb,amsfonts, tabularx}
\usepackage{gensymb, wasysym} 
\usepackage{graphicx, xcolor, subcaption, soul}
\usepackage{nameref, zref-xr}
\usepackage{algorithmic}
\usepackage{comment}
\usepackage[]{mdframed}
\zxrsetup{toltxlabel}
\usepackage{microtype}
\usepackage[nolist]{acronym}
\definecolor{myBlue}{HTML}{03456A} 
\definecolor{Gray}{RGB}{230,230,230}
\usepackage{hyperref}
\hypersetup{
    colorlinks=true,
    linkcolor=myBlue,
    citecolor=myBlue,
    filecolor=myBlue,      
    urlcolor=myBlue,
    pdftitle={main},
    pdfpagemode=FullScreen,
    }
\usepackage[nameinlink]{cleveref}
\crefname{figure}{Fig.}{Figs.}
\Crefname{figure}{Fig.}{Figs.}
\crefname{table}{Table}{Tables}
\Crefname{table}{Table}{Tables}
\usepackage{siunitx}
\usepackage{threeparttable}

\usepackage{dsfont}

\newcommand\etal[0]{\textit{et\,al.\ }}
\newcommand\mailto[1]{\href{mailto:#1}{#1}}

\let\oldcite\cite
\renewcommand*\cite[1]{\,\oldcite{#1}}
\def\BibTeX{{\rm B\kern-.05em{\sc i\kern-.025em b}\kern-.08em
    T\kern-.1667em\lower.7ex\hbox{E}\kern-.125emX}}

%% file: acronym.tex
\begin{acronym}
    \acro{DMS}{Distributed Manipulator System}
    \acro{DoF}{Degrees of Freedom}
    \acro{ADC}{Analogue to Digital Converter}
    \acro{TPU}{Thermoplastic Polyurethane}
    \acro{PETG}{Polyethylene Terephthalate Glycol}
    \acro{RL}{Reinforcement Learning}
    \acro{PPO}{Proximal Policy Optimization}
    \acro{DCT}{Discrete Cosine Transform}
    \acro{C2C}{Centre-to-Centre Spacing}
\end{acronym}

%% file: Sections/0_Abstract.tex
\begin{abstract}
    Distributed manipulator systems manipulate objects through the coordinated motion of many actuators. However, an object must be supported by several actuators at once, so the centre-to-centre actuator spacing imposes a hard lower bound on manipulable object size. We remove this bound by coupling the end-effectors of an $\numproduct{8 x 8}$ array of three degrees-of-freedom delta robots with a stretchable fabric, turning 64 discrete contacts into a continuous surface capable of manipulating objects smaller than the actuator spacing. Viewing the array as a displacement field over that surface, we investigate local quasi-static and cyclic fields as manipulation primitives. These primitives can be applied globally across the array or locally confined around each tracked object to independently manipulate several objects in parallel. We then train a policy acting on low-order discrete cosine transform coefficients: at equal action dimension, commanding a nineteen-delta neighbourhood halves the placement error of commanding the whole array. The policy transfers to hardware without adaptation at \SI{76}{\percent} success. The platform manipulates objects from $\qtyrange{15}{90}{\mm}$, a six-fold range spanning both sides of the actuator spacing, $\SI{43.3}{\mm}$, on a single surface.
\end{abstract}

%% file: Sections/1_Introduction.tex
\section{Introduction}
\label{sec:intro}

\acp{DMS} manipulate objects not with a single end-effector but through the coordinated motion of many actuators arranged in an array. %
This motion is stabilised through many points of contact occurring simultaneously. Such systems have shown their utility manipulating a range of object morphologies, including those that may prove challenging for traditional gripper based approaches. Their high bandwidth offers the potential for parallel manipulation capabilities, and actuator redundancy gives them inherent robustness to failure\cite{bohringerDistributedManipulation2000}.

However, a limitation shared by discrete actuator arrays is that the object
must rest on several actuators simultaneously; at least three
non-collinear supporting actuators \cite{luntzDiscretenessIssuesActuator2000}. It follows that the \ac{C2C} between actuators sets a lower bound on object size: anything smaller falls between the actuators. The range of manipulable objects is therefore bounded from below by the hardware; a denser array can only lower this floor, not remove it. Extending distributed manipulation to very small objects without a proportionally dense array requires a way to provide continuous contact, not only at discrete points.

\begin{figure}[t]
    \centering
    \includegraphics[width=\linewidth, trim ={50 850 50 150}, clip]{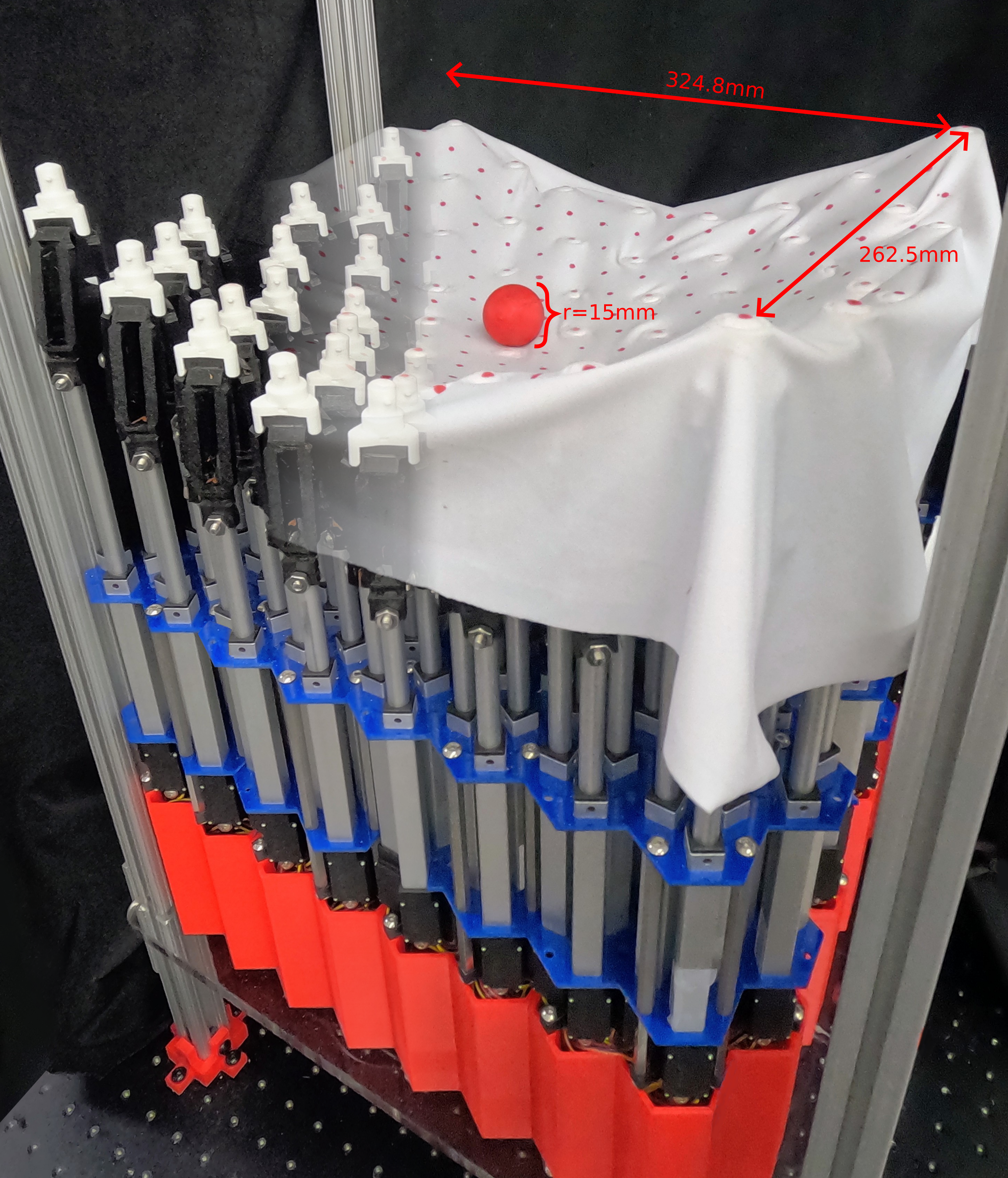}
    \caption{An $8\times8$ array of $3$-DoF delta robots whose end-effectors are coupled by a stretchable fabric, turning $64$ discrete contacts into a continuous surface. Objects both larger and smaller than the inter-actuator spacing are manipulated by shaping the surface.}
    \label{fig:system}
\end{figure}

In this paper, we remove the constraint imposed by the \ac{C2C} by interconnecting the end-effectors of an actuator array with a stretchable material (Fig.~\ref{fig:system}). The end-effector positions control the shape of the continuous surface formed, so objects smaller than the \ac{C2C}, can be manipulated by utilizing the local slope, curvature, and strain, here demonstrated down to $\SI{15}{\mm}$ on an array with a $\SI{43.3}{\mm}$ \ac{C2C}. Larger objects can still be  driven by the coordinated motion of the actuators beneath them, with the membrane mediating contact in-between. A single platform therefore spans both regimes. As the actuators now control a continuous surface, it becomes natural to reframe actuation as a field over the surface: quasi-static fields deform the surface locally, moving an object either under gravity or ejection from the strained material; cyclic fields impart velocity directly by producing travelling waves and gaits.

We demonstrate these ideas on an $8\times8$ array of $3$-\ac{DoF} delta robots, $192$~\ac{DoF} in total, coupled together by an elastic membrane. Here, we make the following contributions:
\begin{itemize}
  \item Characterisation of the connected workspace of the coupled array including the strain induced in the material as the deltas move through it.
  \item Quasi-static fields that use the first order in-plane and second order out-of-plane linearised surface kinematics to translate objects, and cyclic fields producing travelling waves and gaits for object manipulation.
  \item Closed-loop manipulation across object scales: objects smaller than the actuator \ac{C2C} are routed via local deformation of the surface with dynamic re-planning and collision avoidance, orchestrated by a high-level planner. Larger objects are translated and rotated independently by localised cyclic fields.
  \item A trained policy acting on low-order coefficients of the discrete cosine transform of the delta positions. The model translates objects spanning a range from $\qtyrange{30}{90}{\mm}$ in characteristic length and transfers from simulation to hardware without adaptation.
\end{itemize}

%% file: Sections/2_RelatedWork.tex
\section{Related Work}

\subsection{Distributed Manipulation Systems}

\ac{DMS}, systems composed of a distributed array of actuators, have been researched for their object manipulation capabilities, including arrays of pistons~\cite{xueArrayBotReinforcementLearning2024, ingleSoftManipulationSurface2025, follmerInFORMDynamicPhysical2013}, wheels~\cite{uriarteMethodeZurBewertung2022}, parallel mechanisms~\cite{patilLinearDeltaArrays2023, dacreScalableLowDensityDistributed2026}, and cilia-like MEMS devices~\cite{bohringerTheoryManipulationControl1994}, differing in scale and in the \ac{DoF} available in and out of plane. 
Most systems use a single \ac{DoF} per element; however parallel mechanisms have been investigated for their utility in providing independent in-plane and out-of-plane motion \cite{patilLinearDeltaArrays2023, dacreScalableLowDensityDistributed2026}.
Despite the variety, most share a common architecture: a dense array manipulating objects substantially larger than the \ac{C2C} between adjacent actuators. This is required for the continuous vector-field model to hold \cite{bohringerDistributedManipulation2000}; as object size approaches the \ac{C2C}, discrete contact effects dominate  \cite{luntzDiscretenessIssuesActuator2000}, and the object can no longer maintain the multiple contacts needed for stable support.

\subsection{Continuous Surface-Based Manipulation}

Rather than grasping objects, surface-based platforms manipulate objects by dynamically deforming a continuous surface on which the objects rest, using motions that leverage friction, gravity, and induced impulse to reposition one or multiple objects simultaneously \cite{ingleSoftManipulationSurface2025, festoWaveHandling2013}. These systems enable non-prehensile manipulation, making them well suited to objects that are soft, fragile, or of irregular geometry. Surface-based systems are also not restricted by the typical \ac{DMS} actuator \ac{C2C} constraints, these platforms can transport smaller objects owing to continuous surface contact. 

Existing surface-based manipulators can be categorised by the tension of the contact interface. In taut interfaces~\cite{festoWaveHandling2013}, the surface is shaped by elastic deformation of the material, so its geometry between contacts follows the actuator displacements. In slack interfaces~\cite{ingleSoftManipulationSurface2025, dacreScalableLowDensityDistributed2026}, the material hangs under gravity between sparse fixings, forming a catenary whose shape is set by the fixing points and the material's weight; this conforms to cradle irregular objects, but the actuators can only directly control the fixings and must shape the span between them at a distance. In both cases, the choice between elastic deformation and slack is fixed during design by material and mounting. Where actuation is purely prismatic~\cite{festoWaveHandling2013, ingleSoftManipulationSurface2025}, in-plane strain is coupled to out-of-plane displacement and cannot be commanded independently; Dacre~\etal~\cite{dacreScalableLowDensityDistributed2026} provide three DoF per element via Canfield joints, but use an inextensible sheet. Our system combines three \ac{DoF} per node with a stretchable membrane held at neutral strain, so local strain becomes a bidirectional command: a region can be relaxed into the slack regime to cradle objects smaller than the \ac{C2C}, or stretched elastically to shape and drive them.

\subsection{Learned Control}

ArrayBot \cite{xueArrayBotReinforcementLearning2024} learns policies for a rigid array of pistons by acting on low-order \ac{DCT} modes. This keeps the  action space tractable. We adopt the same spectral action space on a compliant, membrane-coupled surface, where the smoothness of the commanded field is also a physical constraint of the material.

%% file: Sections/3_MaterialsAndMethods.tex
\section{Materials and Methods}
 We use a system consisting of $64$ delta robots, arranged in a hexagonal $8\times8$ array.
The \ac{C2C} between each delta's base is \SI{43.3}{\milli\meter} \cite{patilLinearDeltaArrays2023}.

\begin{figure}
    \centering
    \includegraphics[width=\linewidth]{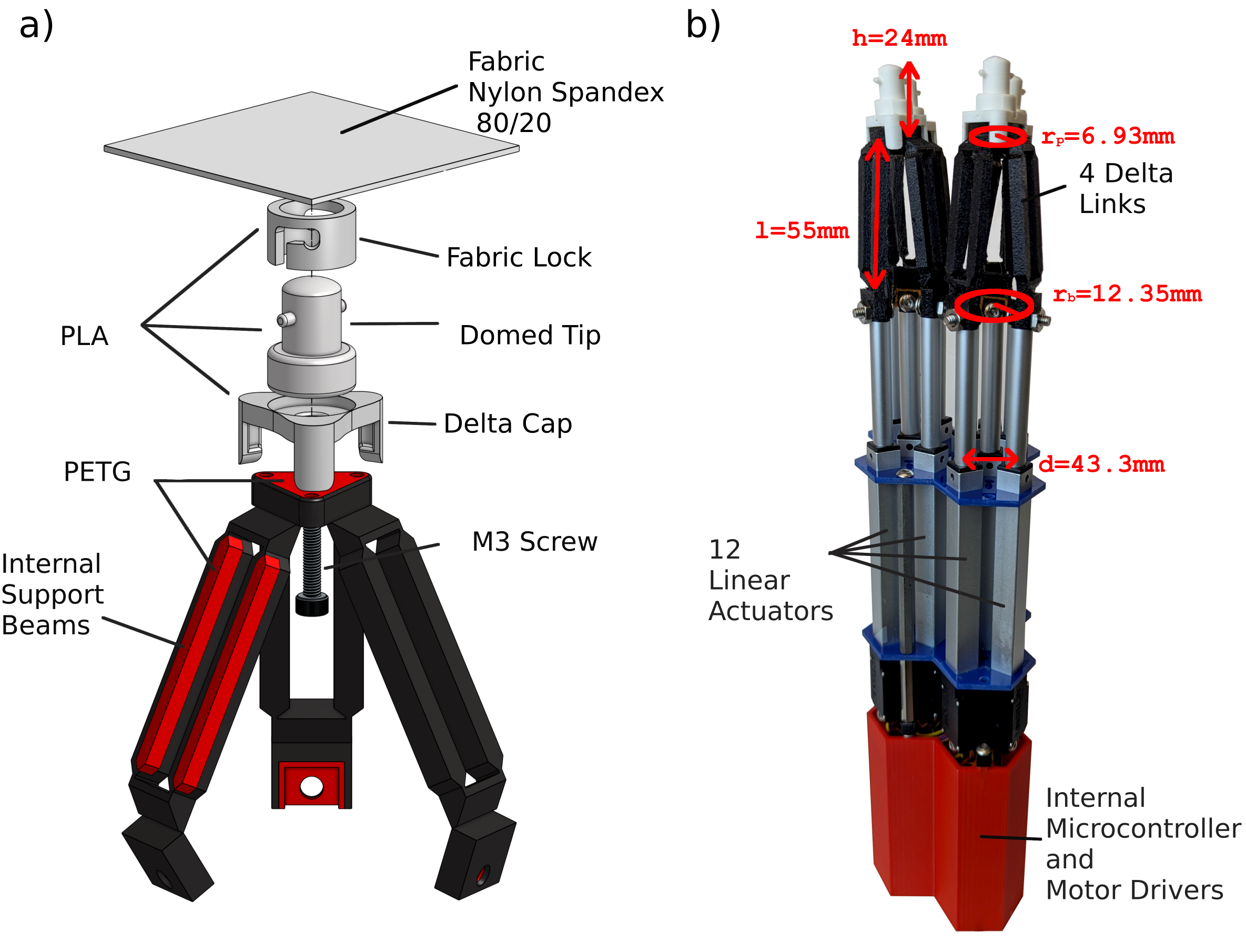}
    \caption{a) Exploded view of a delta and quick release fabric attachment mechanism. b) A $2\times2$ unit. The array is made up of $16$ of these units. Kinematic parameters (\Cref{sec:single_workspace}) given in red.}
     \label{fig:delta_unit}
\end{figure}

The delta mechanisms are a combination of three parallelogram linkage structures made from \ac{TPU} and \ac{PETG} (\cref{fig:delta_unit}(a)), 3D-printed using a dual-extrusion printer (Bambu Lab H2D). \ac{TPU} is used for its high compliance, allowing the delta to bend along living hinges printed in the structure. Internal \ac{PETG} beams reinforce the parallel linkages against bending. %

Each delta is controlled by three linear actuators (Actuonix L12-100-50-12-P), with an effective \SI{100}{\mm} stroke length and inbuilt potentiometers for positional feedback. %

The array comprises 16 modular 2 × 2 units each with four deltas (12 actuators)\cite{patilLinearDeltaArrays2023}(\cref{fig:delta_unit}(b)).
Each unit has its own microcontroller (Adafruit Feather M0), three motor drivers (DC Motor Featherwing), \ac{ADC}, attached to a custom PCB.

A fabric membrane is attached to the deltas via a quick-release twist-lock mechanism, designed for ease of access and to allow the fabric to be replaced as needed. The inter connective material  is an $80/20$ Nylon Spandex blend with a thickness \SI{0.4}{\mm}, density \SI{200}{\gram \meter^-2} and a manufacturer supplied maximum achievable strain of $70$\%. This forms a continuous surface with a  $263 \times$\SI{325}{\mm} footprint.

For object tracking, we utilise an array of four cameras (MOKOSE, Logitech) positioned to view the surface from different vantage points. The cameras capture in 1080p at 30 FPS. Points of interest are tracked by an intensity-weighted image moments tracker built from the OpenCV library \cite{bradskiOpencvLibrary2000} and localised with AprilTags.

%% file: Sections/4_WorkspaceAnalysis.tex
\section{Workspace Analysis}
\subsection{Single-module workspace}
\label{sec:single_workspace}

Each delta is driven by three parallel prismatic actuators \cite{patilLinearDeltaArrays2023}, so the end effector stays parallel to the base and each link acts as a scalar distance constraint. Actuator $i$ sits at azimuth $\theta_i \in \{\frac{\pi}{2}, \frac{7\pi}{6}, \frac{11\pi}{6}\}$ about the base centroid, with an in-plane direction $\hat{\mathbf{u}}_i$; base and platform circumradii are $r_b = \SI{12.35}{\mm}$ and $r_p = \SI{6.93}{\mm}$, link length $\ell = \SI{55}{\mm}$, and the tip sits $h = \SI{24}{\mm}$ above the platform (\cref{fig:delta_unit}). Each lateral section of the workspace $\mathcal{W}$ is thus the intersection of three discs of radius $\ell$ centred at $(r_b - r_p)\hat{\mathbf{u}}_i$, truncated in $z$ by the \SI{100}{\mm} actuator stroke. $\mathcal{W}$ is therefore $C_{3v}$ symmetric, with mirror planes aligned with the actuator azimuths. %

\subsection{Connected workspace}
\label{sec:connected_workspace}

In the membrane-connected array, each delta can strain the membrane according to its pose relative to its neighbours. The tips sit on a regular triangular lattice of pitch $d = \SI{43.3}{\mm}$, the \ac{C2C}. This gives each interior delta six neighbours. The membrane is mounted flat at its unstretched length, $\varepsilon = 0$; any residual pretension is negligible against commanded strains. Modelling the span between adjacent tips as a straight line and neglecting out-of-plane deflection and shear, two neighbours separated by lattice vector $\mathbf{a}$, $\lVert\mathbf{a}\rVert = d$, with tip displacements from neutral $\mathbf{x}_a, \mathbf{x}_b \in \mathcal{W}$ and relative displacement $\boldsymbol{\delta} = \mathbf{x}_b - \mathbf{x}_a$, span $s = \lVert\mathbf{a} + \boldsymbol{\delta}\rVert$ and carry engineering strain $\varepsilon = (s - d)/d$. The connected workspace is the set of pose pairs with $\varepsilon \le \varepsilon_{\max}$, equivalently $s \le s_{\max}$, where $ s_{\max} = (1 + \varepsilon_{\max})\,d$.

For a workspace discretised into $N$ points, enumerating all $6 N^2$ neighbour pairs for an interior delta is unnecessary. The span depends on a pair only through $\boldsymbol{\delta}$, so what is needed is the multiplicity of each displacement, which on a grid with occupancy indicator $A$ is the autocorrelation
\begin{equation}
  C(\boldsymbol{\delta}) = \sum_{\mathbf{x}} A(\mathbf{x})A(\mathbf{x}+\boldsymbol{\delta})
  = \mathcal{F}^{-1}\!\left\{\lvert\mathcal{F}\{A\}\rvert^2\right\}.
  \label{eq:wk}
\end{equation}

\begin{figure}
    \centering
    \includegraphics[width=\linewidth]{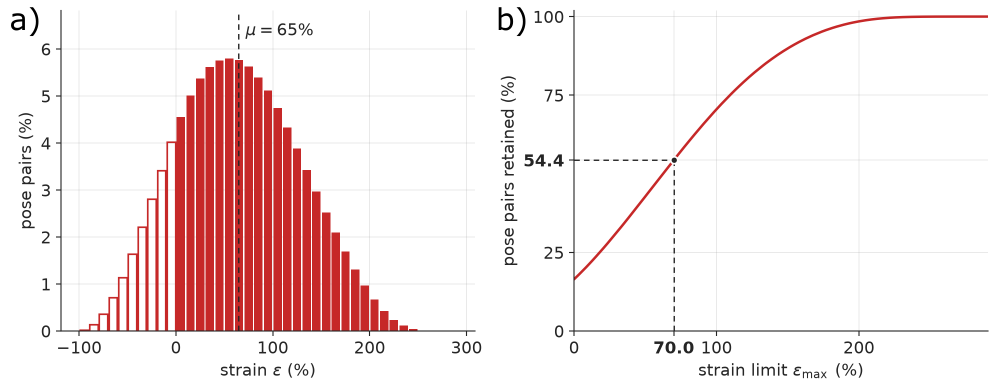}
    \caption{Strain over all $4.9\times10^{11}$ pose pairs of an adjacent delta pair, from the workspace autocorrelation at \qty{1}{\mm} resolution. (a) Histogram in \qty{10}{\percent} bands. (b) Admissible fraction of the joint workspace under $\varepsilon\le\varepsilon_{\max}$, with the \qty{70}{\percent} limit of the membrane used marked.}
    \label{fig:strain_histogram_and_curve}
\end{figure}

By the convolution theorem \cite{smithMathematicsDiscreteFourier2007}, this calculation costs $O(M\log M)$ in grid cells rather than $O(N^2)$ in pairs. The six neighbour directions fall into two $C_3$ orbits,%
$\frac{\pi}{6}, \frac{5\pi}{6}, \frac{3\pi}{2}$ and $\frac{\pi}{2}, \frac{7\pi}{6}, \frac{11\pi}{6}$, which the $C_{3v}$ symmetry of $W$ alone does not relate. Reversing an edge, however, negates $\boldsymbol{\delta}$ without changing the span, so $C(-\boldsymbol{\delta})=C(\boldsymbol{\delta})$ for any $W$; inversion relates the two orbits and all six neighbours share one span-length distribution. 
The distribution is  obtained by binning $\lVert\mathbf{a}+\boldsymbol{\delta}\rVert$ against weights $C(\boldsymbol{\delta})$ for a single neighbour. At the $\epsilon_{\max} = 0.7$ strain limit of the fabric used here, $54.4$\% of pose pairs remain, as shown in \cref{fig:strain_histogram_and_curve}. %

The model assumes uniaxial loading along each edge. Within it, adjacent tips reach a maximum height differential of $\SI{59.1}{\mm}$, a local gradient of $\SI{0.94}{\radian}$.

However, pairwise admissibility does not compose: for a delta and its six neighbours, 21 \ac{DoF} and twelve spans, uniform Monte Carlo over $4\times10^{7}$ configurations finds \SI{0.42}{\percent} of joint space admissible at $\varepsilon_{\max}=0.7$, against $0.544^{12}=\SI{0.07}{\percent}$ if spans were independent; spans sharing a delta are positively correlated. Admissible configurations therefore form a small, coherent subset of joint space, which is why the array is commanded through fields (\cref{sec:fields}) and low-order modes (\cref{sec:rl}).

\begin{figure}
    \centering
    \includegraphics[width=0.65\linewidth, trim={0 0 0 50}, clip]{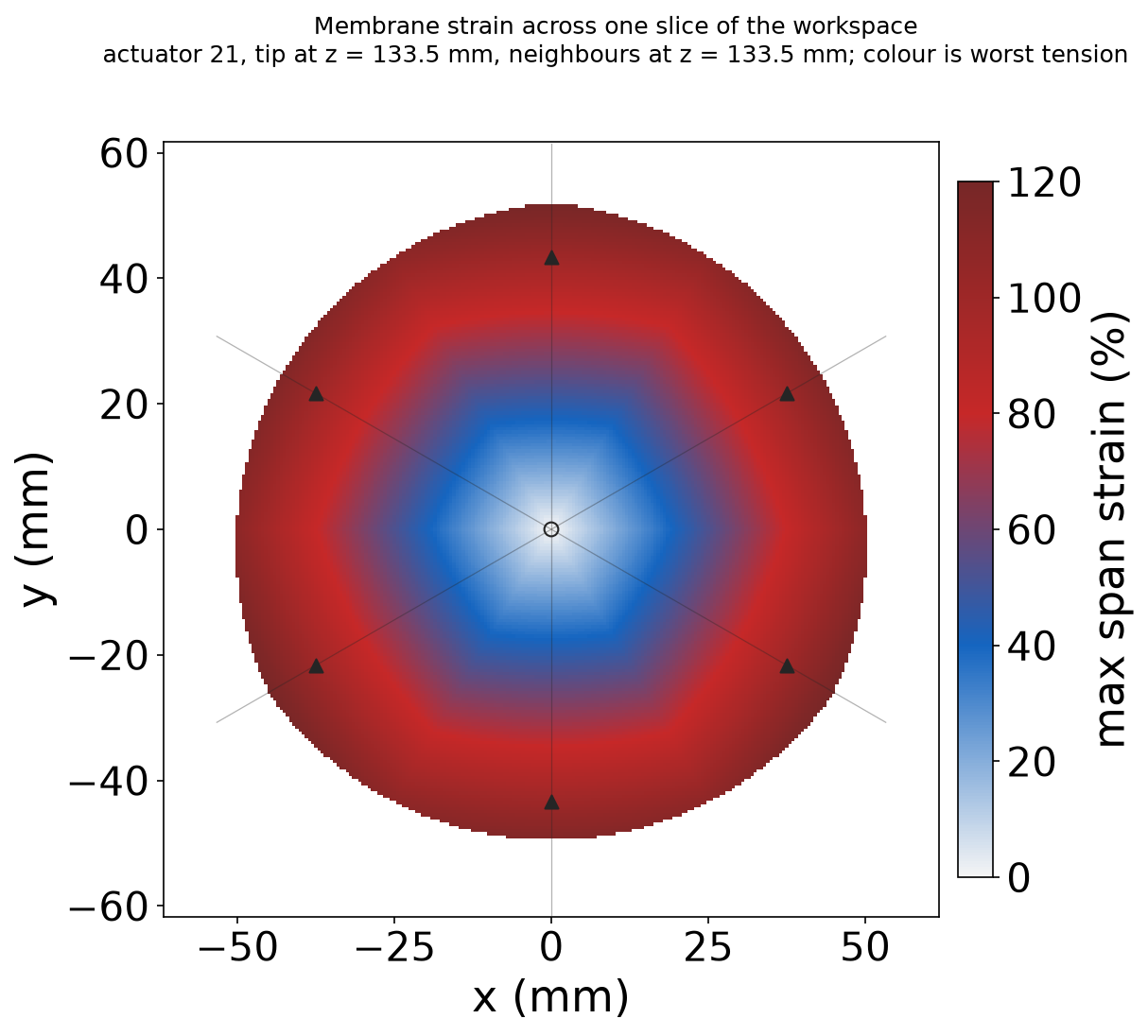}
    \caption{Single delta Workspace slice at $z=\qty{133}{\mm}$, coloured by the maximum membrane strain between one delta and its six neighbours, all held at their neutral poses (black triangles). %
    }
    \label{fig:strain_workspace}
\end{figure}

%% file: Sections/5_FieldsAndPrimatives.tex
\section{Surface Fields and Local Primitives}
\label{sec:fields}

The membrane surface is set by the tip positions of the deltas, so we
command it through a spatially sampled displacement field $\mathbf{u}(\mathbf{p}_i, t) = (\mathbf{v}, w)$: the in-plane and out-of-plane displacement of delta $i$ from neutral, as a function of its lattice position $\mathbf{p}_i$ and a global clock $t$. Because all deltas share a common orientation, a field specified over the surface is directly a valid per-delta command. 

To describe a field, we denote $\hat{\mathbf{d}}$ as a freely specified in-plane unit direction, used to orient the field. For fields with a centre $\mathbf{p}_c$ we write $\mathbf{r} = \mathbf{p}_i - \mathbf{p}_c = (X, Y)$.

The same primitives serve both open and closed-loop control, differing only in how these parameters are set. Under open-loop control, the parameters are fixed in advance and the field is applied globally across the whole array. Under
closed-loop control, the parameters can be updated at each time step from the tracked object pose, obtained by triangulation with the four-camera rig and transformed to the array frame via static AprilTags. For example $\mathbf{p}_c$ could follow or lead the object, $\hat{\mathbf{d}}$ may be used to direct toward its goal.
Applied to only the subset of deltas surrounding an object, a field can be applied locally. In such a way, several objects can be manipulated independently and concurrently on a single surface.

\subsection{Field Primitives}
\label{sec:fields:primitives}

\textbf{Quasi-static fields.} We define twelve quasi-static shape fields spanning the linearised kinematics of the surface to first order in-plane and second order out-of-plane (\cref{tab:static_fields}): six rigid-body modes (three translations, two tilts, one in-plane rotation) and six deformation modes, comprising three membrane strains (dilation and two shears) and three bending curvatures (a parabola and two saddles).

By controlling the membrane's curvature and slope, these fields can be used for manipulation: tilt drives sliding or rolling under gravity, the parabola collects or disperses rolling objects about $\mathbf{p}_c$, and the in-plane strains reposition and reorient objects
resting on the surface.

\begin{table}[t]
\caption{The twelve quasi-static shape modes, each an in-plane displacement $\mathbf{v}$ or out-of-plane displacement $w$ of a delta at $\mathbf{r}=(X,Y)$ from the field centre. Amplitudes: heave $h$ [m], gradient $g$, curvature $\kappa$ [m$^{-1}$], offset $c$ [m], dilation $\beta$, rotation $\Omega$ [rad], shear $\gamma$. Two-DoF modes are listed as a basis pair; their combination sets magnitude and orientation.}
\label{tab:static_fields}
\centering
\begin{tabular}{@{}llc@{}}
\toprule
Mode & Basis & DoF \\
\midrule
\multicolumn{3}{@{}l}{\emph{Out-of-plane} ($\mathbf{v}=\mathbf{0}$): $w=$}\\
\quad Heave       & $h$                                                  & 1 \\
\quad Tilt        & $gX \;\mid\; gY$                                      & 2 \\
\quad Parabola    & $\tfrac{1}{2}\kappa\,(X^{2}+Y^{2})$                  & 1 \\
\quad Saddle      & $\tfrac{1}{2}\kappa\,(X^{2}-Y^{2}) \;\mid\; \kappa XY$ & 2 \\
\midrule
\multicolumn{3}{@{}l}{\emph{In-plane} ($w=0$): $\mathbf{v}=$}\\
\quad Translation & $c\,(1,0) \;\mid\; c\,(0,1)$                          & 2 \\
\quad Dilation    & $\beta\,(X,Y)$                                  & 1 \\
\quad Rotation    & $\Omega\,(-Y,X)$                                      & 1 \\
\quad Shear       & $\gamma\,(X,-Y) \;\mid\; \gamma\,(Y,X)$               & 2 \\
\midrule
\textbf{Total}    &                                                      & \textbf{12} \\
\bottomrule
\end{tabular}
\end{table}

\textbf{Cyclic fields.} A cyclic field drives each tip around a closed stroke $\sigma$, returning to its initial pose each cycle. Transport arises from the asymmetry between the loaded and unloaded portions of the stroke. These are demonstrated on delta arrays by Patil \etal as gaits 
\cite{patilLinearDeltaArrays2023}.

Three parameter sets specify cyclic fields. The \emph{stroke}, here common to all deltas, fixing per-cycle displacement and duty ratio. The \emph{direction field} $\hat{\mathbf{d}}(\mathbf{r})$ orients each stroke and is drawn from the same in-plane basis as \cref{tab:static_fields} - uniform, radial, tangential, and shear - so the transport modes inherit the completeness of the quasi-static set. The \emph{phase field} $\varphi(\mathbf{r})$ assigns a temporal offset, with delta $i$ executing $\mathbf{\sigma}(\omega t + \varphi(\mathbf{p}_i))$; this determines how travelling waves are sequenced across the surface. %

The membrane acts as a physical reconstruction filter: the deltas sample the wave at lattice sites and the membrane interpolates between them, presenting a continuous surface rather than a sequence of discrete steps. Objects smaller than the \ac{C2C} are therefore transported without loss of contact; we return to the implications of reconstruction in \Cref{sec:discussion}.

For a travelling wave, $\phi(\mathbf{r}) = \mathbf{k}\cdot\mathbf{r} \bmod 2\pi$ with wave vector $\mathbf{k}$, heading $\theta = \operatorname{atan2}(k_y, k_x)$ and wavelength $\lambda = 2\pi/\|\mathbf{k}\|$. Independent actuation of each delta admits a continuously valued $\phi(\mathbf{r})$, so the wave may take any heading; a gait built from $N$ discrete phases~\cite{patilLinearDeltaArrays2023} is exact only at headings where the phase advance between neighbours is a multiple of $2\pi/N$.

\begin{figure}
    \centering
    \includegraphics[width=0.65\linewidth, trim={0 0 0 0}, clip]{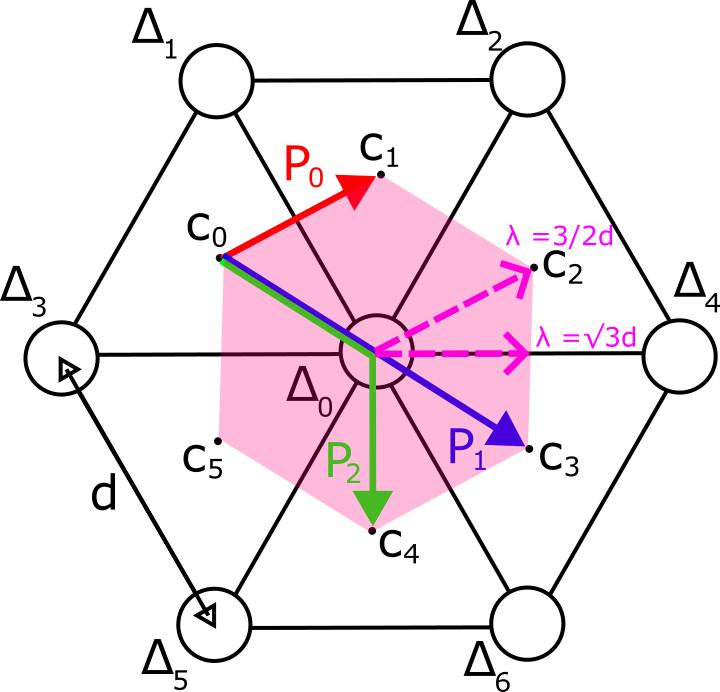}
    \caption{A central delta and its ring-1 neighbourhood, $R_1$. Purple marks the unit cell of the reciprocal lattice and $\lambda_{\min}$ in the principal directions. Inter-cell primitives - edge ($P_0$, red), node ($P_1$, blue) and deflect ($P_2$, green) - transfer an object between cells, $C_n$, triangular regions of fabric between adjacent deltas.}
    \label{fig:cell_primitives}
\end{figure}

The \ac{C2C}, $d$, still bounds the reproducible wavelength. Since the wave is rendered only where deltas exist, a $\mathbf{k}$ advancing the phase by $2\pi$ between neighbours moves them in unison and is indistinguishable from infinite wavelength; shortening the wave further aliases to a longer wave at a different heading rather than finer structure. Over all headings the admissible $\mathbf{k}$ therefore lie within the unit cell of the reciprocal lattice~\cite{petersenSamplingReconstructionWavenumberlimited1962,mersereauProcessingHexagonallySampled1979}, a hexagon (\cref{fig:cell_primitives}) whose boundary corresponds to a $\pi$ advance along a lattice direction, adjacent deltas in antiphase; its boundary distance sets a heading-dependent minimum wavelength, $\lambda_{\min}$:

\begin{equation}
  \lambda_{\min}^{\theta} = \sqrt{3}\,d
  \max_{i\in\{0,1,2\}}\bigl|\cos\bigl(\theta - \tfrac{i\pi}{3}\bigr)\bigr|
  \in \left[\tfrac{3}{2}d,\ \sqrt{3}\,d\right].
  \label{eq:lambda_min}
\end{equation}

\begin{figure}
    \centering
    \includegraphics[width=0.9\linewidth, trim={25 170 188 85}, clip]{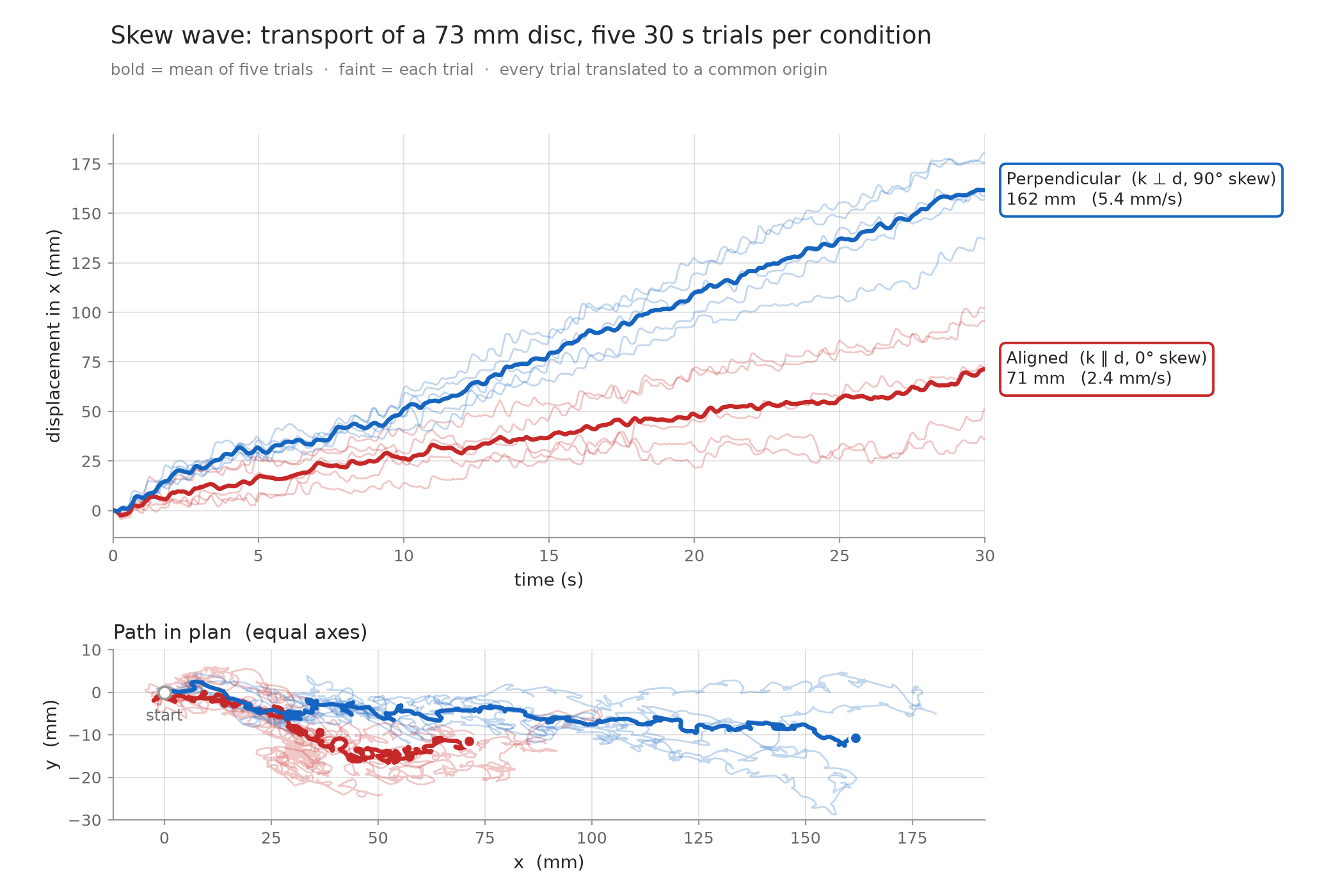}
    \caption{Translation of a disc ($\diameter = \qty{70}\mm, \frac{3d}{2} < \diameter < \sqrt{3} d$) across the array using an elliptical gait with $\hat{\mathbf{d}}=[1,0]$, averaged (bold line) over five trials (faint). Red: wave vector $\mathbf{k}$ aligned with $\hat{\mathbf{d}}$; blue: $\mathbf{k}$ perpendicular, showing a greater translational velocity. Each uses the shortest wavelength available at that heading, $\lambda_{\min}^{\theta}$.}
    \label{fig:perpendicular_translation}
\end{figure}

The three \ac{DoF} decouple where objects travel from how support is sequenced. A vertically actuated element encloses no orbit and can only drag an object along the propagating wave, fixing transport to the phase gradient. A closed stroke instead drives the object along $\hat{\mathbf{d}}(\mathbf{r})$ during contact, leaving $\mathbf{k}$ free to serve support alone. \Cref{fig:perpendicular_translation} shows how this can be utilised to efficiently gait an object in a direction in which $\lambda_{\min}^{\theta}$ would be too long to provide stable support, if $\mathbf{k}$ and $\hat{\mathbf{d}}(\mathbf{r})$ were forced to be aligned.

\subsection{Cell and Ring Local Primitives}
\label{sec:local_primitives}
In their neutral positions, the tips form an equilateral triangular lattice of \ac{C2C} $d$ (\cref{fig:cell_primitives}). Treating the tips as nodes of this lattice, the triangular regions of fabric between three mutually adjacent tips are \emph{cells}, whose local shape is set primarily by these three deltas. An object small enough to sit within a cell is supported by the membrane alone and is manipulated by shaping that cell. A larger object spans several cells and is supported by the membrane and the tips beneath it. A local surface enclosed by the neighbourhood of deltas centred around the tip nearest the object, which we define as a \emph{ring}. 
Each case has its own family of primitives, and a planner routes objects through the corresponding graph: cell to cell or ring to ring.

\begin{figure}
    \centering
    \includegraphics[width=\linewidth]{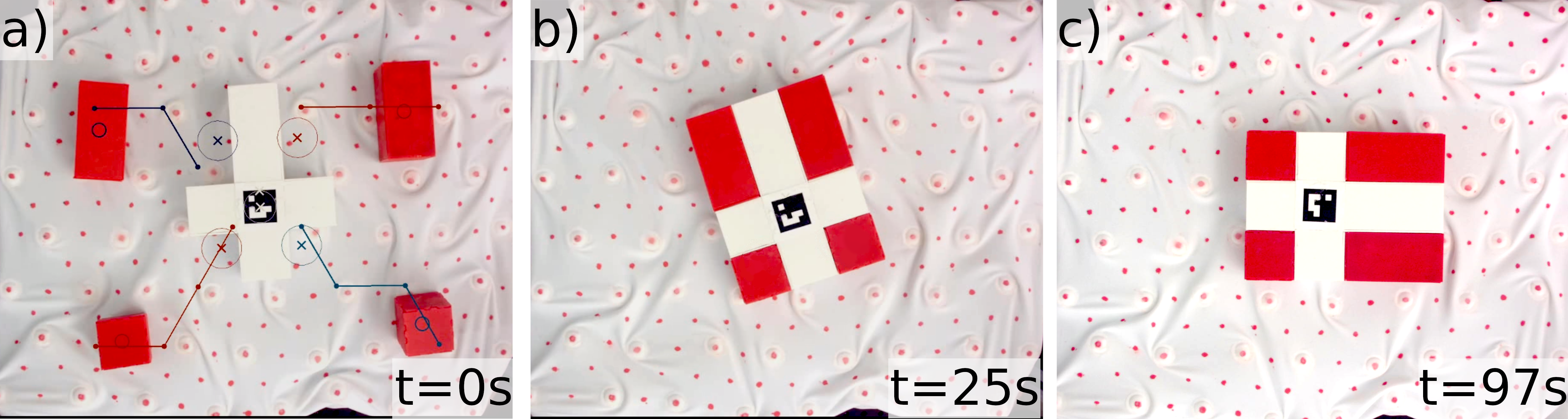}
    \caption{Assembly and orientation of magnetic blocks, localised by AprilTag. (a)~Blocks assembled with the $R_1$ tilt primitive. (b) A local cyclic wave rotates and centres the assembly. (c) Final desired pose.}
    \label{fig:flag_assembly}
\end{figure}

\textbf{Ring primitives.} Let $R_n(i)$ denote the ring-$n$ patch about tip $i$: every tip within $n$ lattice hops of $i$, centre included, so $|R_1| = 7$ and $|R_2| = 19$, spanning $2dn$ and truncated at the array boundary. A ring primitive applies one of the out-of-plane quasi-static modes of Table~\ref{tab:static_fields} or a cyclic field to the patch: tilt translates an object toward a neighbouring node under gravity (Fig.~\ref{fig:flag_assembly}a), and the parabola collects or disperses objects about the centre. Cyclic fields confined to $R_n$ 
translate or rotate objects too large to be moved by tilt alone
(\cref{fig:flag_assembly}b).

\textbf{Cell primitives.} Ring primitives use only the out-of-plane kinematics of the surface. Objects smaller than the \ac{C2C} sit within a cell, where the bounding deltas and their neighbours can shape the fabric using all three \ac{DoF}. We define three cell primitives that transfer an object from a source cell $C_0$ to a neighbour (\cref{fig:cell_primitives}, \cref{tab:cell_primitives}): \emph{edge}, $P_0$, crossing a shared edge into $C_1$; \emph{node}, $P_1$, crossing over a tip into the opposite cell $C_2$; and \emph{deflect},  turning across a tip into $C_3$. Each is a set of commanded offsets for the deltas around the source cell, tuned once and applied by symmetry to every cell and heading. Deflection has two mirror-image variants, and measuring $\hat{\mathbf{c}}$ toward $\Delta_0$ lets one tuning serve both. These primitives require the object to fit within a cell; the inscribed circle has diameter $d/\sqrt{3} \approx 25$\,mm, but in practice, transfer remained reliable up to approximately $40$\,mm, as some out of plane overhang beyond the cell boundary was not detrimental. Utilizing the 3-\ac{DoF} of each actuator allows for slack to be created to capture objects, or elastic strain to be created to eject objects. Each primitive is a single target pose. As the deltas move to it, the strained fabric ejects the object from the source cell and it descends the resulting gradient into the target; the pose is held until tracking confirms traversal.

\begin{table*}[t]
\centering
\small
\setlength{\tabcolsep}{5pt}
\caption{Cell primitives of \cref{fig:cell_primitives}, all moving from $C_0$. Each row is a delta's tip offset from neutral (mm), expressed in a frame attached to the primitive: $\hat{a}$ points from $C_0$ toward the target cell, $\hat{z}$ up, and $\hat{c} = \hat{z}\times\hat{a}$ to left of travel. For deflection, $\hat{c}$ is taken toward $\Delta_0$; the mirror variant follows by symmetry.}
\label{tab:cell_primitives}
\begin{tabular}{@{}lrrr@{\hspace{2em}}lrrr@{\hspace{2em}}lrrr@{}}
\toprule
\multicolumn{4}{c}{$P_0$ edge: $C_0\!\to\!C_1$} & \multicolumn{4}{c}{$P_1$ node: $C_0\!\to\!C_2$} & \multicolumn{4}{c}{$P_2$ deflect: $C_0\!\to\!C_3$} \\
\cmidrule(r{2em}){1-4}\cmidrule(r{2em}){5-8}\cmidrule{9-12}
Delta & $\hat{a}$ & $\hat{c}$ & $\hat{z}$ & Delta & $\hat{a}$ & $\hat{c}$ & $\hat{z}$ & Delta & $\hat{a}$ & $\hat{c}$ & $\hat{z}$ \\
\midrule
$\Delta_3$ & $-5$  & $0$   & $+15$ & $\Delta_1$ & $-3$  & $+10$ & $+25$ & $\Delta_1$ & $-5$  & $0$     & $+15$ \\
$\Delta_1$ & $0$   & $+15$ & $-10$ & $\Delta_3$ & $-3$  & $-10$ & $+25$ & $\Delta_3$ & $-10$ & $-17.3$ & $-10$ \\
$\Delta_0$ & $0$   & $-15$ & $-10$ & $\Delta_0$ & $0$   & $0$   & $-20$ & $\Delta_0$ & $0$   & $+20$   & $-10$ \\
$\Delta_2$ & $-10$ & $0$   & $-30$ & $\Delta_4$ & $-10$ & $-10$ & $-25$ & $\Delta_5$ & $+10$ & $-17.3$ & $-20$ \\
           &       &       &       & $\Delta_6$ & $-10$ & $+10$ & $-25$ & $\Delta_6$ & $-10$ & $0$     & $-30$ \\
\bottomrule
\end{tabular}
\end{table*}

\subsection{Routing and Multi-Object Deconfliction}
\label{sec:fields:routing}

For either family of local primitive, an A* planner routes objects from any ring to any ring, or any cell to any cell. The path is verified at every control step against the tracked pose, so a disturbed object is re-planned rather than lost; after three consecutive failed transfers the current edge is penalised and the object re-routed around the obstruction. We evaluated this using a single \SI{15}{\mm} cube over 30 runs on six fixed routes of $25-\SI{185}{\mm}$ spanning the array. Every run reached its goal cell, at a net
speed of $7.6 \pm 2.9$\,mm/s and $28.4 \pm 10.7$\,mm of progress per transfer. $12.3\%$ stalled, leaving the cube in its source cell, but retries succeeded often enough that re-routing was triggered on only $1.5\%$ of transfers.

\begin{figure}
    \centering
    \includegraphics[width=0.75\linewidth]{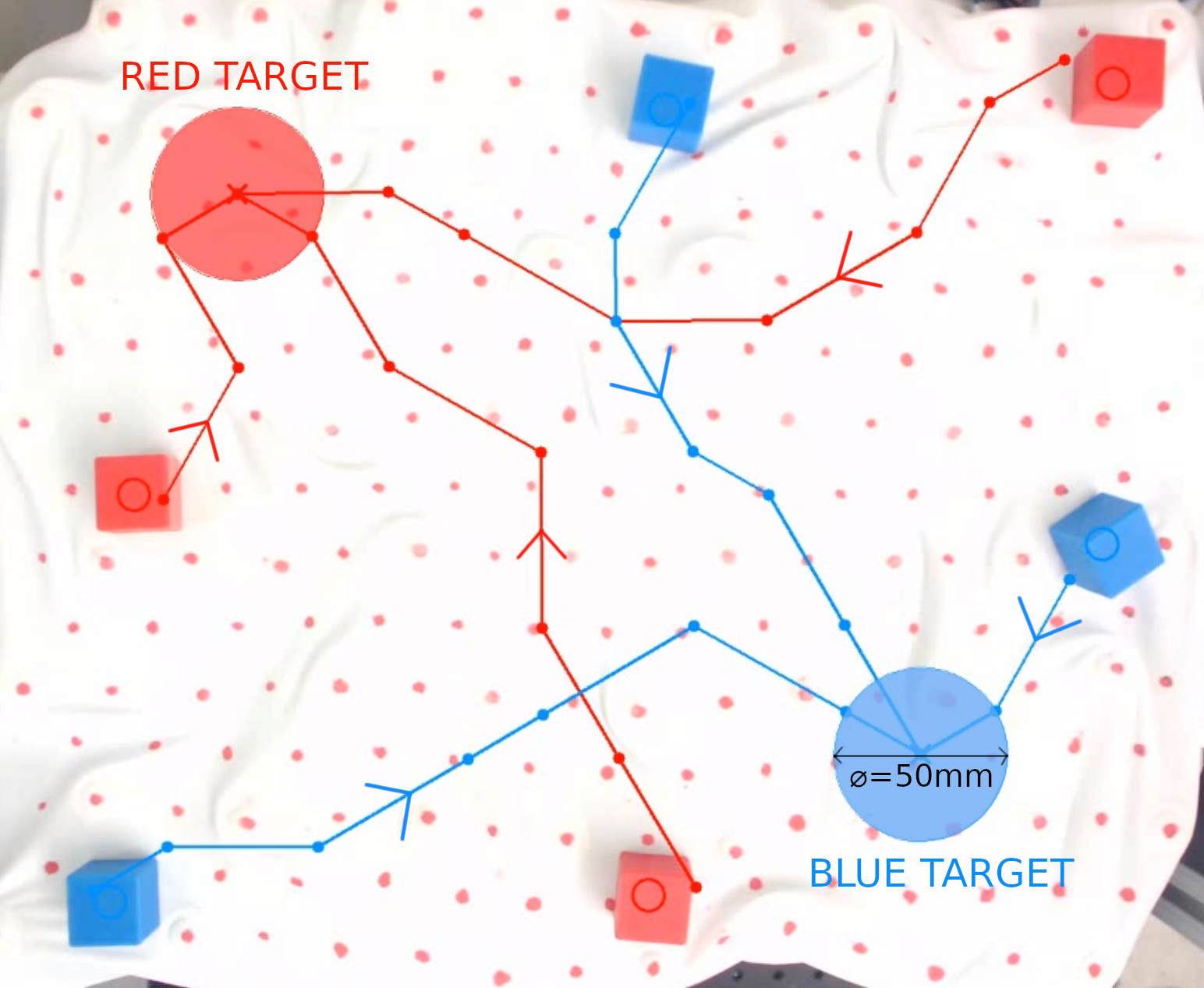}
    \caption{Colour sorting task with \qty{15}{\mm} cubes using cell primitives. Objects route dynamically to colour-matched targets (\diameter = $\SI{50}{\mm}$), traversing the array simultaneously without conflict.}
    \label{fig:colour_Sort}
\end{figure}

Addressability on a coupled surface is not free: a locally issued pose perturbs the fabric beyond the cell it targets, and objects in adjacent cells share bounding deltas by construction. Each object therefore raises the cost of the cells and nodes within its area of influence, so that concurrent routes stay outside one another's influence and objects traverse the array simultaneously without conflict (\cref{fig:colour_Sort}). 

%% file: Sections/7_ReinfocmentLearning.tex
\section{Reinforcement Learning}
\label{sec:rl}

The array has $192$ \ac{DoF}, making it hard to intuit a control policy. Hand-designed primitives make the array tractable by restricting it to configurations whose behaviour is predictable. However, different geometries and object scale need their own primitives and none improves with experience. We instead learn a policy that generalises across scale, requiring only a reward defined on the object state rather than a forward model of the surface.

Explored directly, a $192$-dimensional action space produces large local deformations that destabilise the soft-body solver. We therefore act on low-order \ac{DCT} modes of the tip field, following Xue \etal~\cite{xueArrayBotReinforcementLearning2024}. The action is the $3\times3$ low-order coefficients of a 2D cosine basis fitted to the bounding box of the commanded patch and evaluated at each tip, nine per commanded channel. Discarding high-frequency spatial components keeps the commanded surface smooth. We also investigate the effect of  acting on just a subsection of the array, the ring neighbourhoods $R_n$ (\Cref{sec:local_primitives}) against controlling the array as a whole.

\subsection{Training Environment}

We train a policy using \ac{PPO} in MuJoCo simulator to transport a rigid object to within \SI{20}{\mm} of a planar goal by shaping an elastic membrane over the $64$-tip array. Goals are drawn uniformly over the tip grid's bounding box, inset by $20$\,\% of its span per edge so every goal lies within the array. Objects start at rest at a uniformly sampled position over the same inset region. Episodes run for 100 steps (\qty{20}{\s}) and terminate early when the object leaves the array. The goal height is pinned to the object's resting height and the vertical axis is unscored. Each policy is trained for $1\times10^6$ environment steps.%

The membrane is a MuJoCo flex element at \SI{21.65}{\mm} inter-node distance, attached to actuated tips modelled as servos with first-order lag. Commanded tip positions pass through a closed-form span constraint: the commanded displacement is scaled by $\rho \in [0, 1]$ until no neighbouring pair exceeds $\eta\, s_{\max} = 66.25$\,mm, where $s_{\max}$ is the admissible span  of the connected workspace (\Cref{sec:connected_workspace}) and $\eta = 0.9$ a safety factor, so every output pose lies within the connected workspace. Physics is integrated at \qty{2}{\kilo\hertz} and the policy queried at \qty{5}{\hertz}, within the update rate of the physical system.

Objects are drawn from the EGAD dataset \cite{morrisonEGADEvolvedGrasping2020}, each scaled to a square bounding box of edge $S_e$ sampled log-uniformly over $[30, 150]$\,mm at fixed density. The observation comprises object and goal position and their difference, the third-order \ac{DCT} of the tip state, the projection deficit $1 - \rho$ from the last command, and $S_e$. Each entry is standardised by a running mean and variance estimated online and frozen at evaluation. The reward is:
\begin{equation}
r_t = 10\,(d_{t-1} - d_t) + 0.085\,\mathds{1}_{\mathrm{suc}} - 0.01\,(1 - \rho_t) - 10\,\mathds{1}_{\mathrm{lost}},
\label{eq:reward}
\end{equation}

with $d_t$ the in-plane error [m], $\mathds{1}_{\mathrm{suc}}$ indicating $d_t < 20$\,mm and $\mathds{1}_{\mathrm{lost}}$ the object leaving the array.

Three disturbances are injected in training. Object position carries zero-mean Gaussian noise of \SI{2}{\mm} per axis, redrawn each step, modelling a camera estimate. Each actuator carries a per-episode calibration bias (\SI{2}{\mm} in $x, y$; \SI{0.5}{\mm} in $z$) plus \SI{0.2}{\mm} per-step jitter, applied to the executed command but not the stored target, so the policy acts on a command the actuators do not execute exactly. Observed size carries $3$\,\% per-episode relative error. Training is asymmetric: the critic sees the clean state and drawn biases, the actor only the corrupted stream.

\subsection{Model Comparison}

\begin{table}[t]
\centering
\caption{Test-set performance on holdout by action space and commanded channels. \emph{Suc} is the rate of reaching the \qty{20}{\mm} target; \emph{Med} and \emph{p90} are the median and 90th-percentile final in-plane error. $R_n$ denotes the ring-$n$ neighbourhood.}
\label{tab:training_runs}
\begin{tabular}{llccc}
\toprule
Action space & Ch. & Suc (\%) & Med (mm) & p90 (mm) \\
\midrule
$R_2$, 19 deltas       & $xyz$ & 95.1 & \textbf{3.5}  & \textbf{8.4} \\
$R_2$, 19 deltas       & $z$   & \textbf{95.2} & 4.0  & 10.2 \\
$R_1$, 7 deltas        & $xyz$ & 83.9 & 7.1  & 61.1 \\
Whole array, 64 deltas & $xyz$ & 93.5 & 6.5  & 14.2 \\
\bottomrule
\end{tabular}
\end{table}

\Cref{tab:training_runs} compares action-space extent and commanded channels over 392 held-out episodes (49 shapes, eight each). At equal action dimension, an $R_2$ patch of 19 deltas matched the whole array on success rate and roughly halved its median and 90th-percentile error, indicating that transport is governed by deformation local to the object. Restricting to $R_1$ held the median error but degraded the tail sharply, indicating a catastrophic failure on a subset of objects, possibly consistent with a patch too small to steer objects spanning several \ac{C2C}. Commanding $z$ alone on $R_2$ gave comparable success to $xyz$ with a larger median error, suggesting that vertical shaping dominates long-range transport while lateral tip motion refines placement near the goal.

\subsection{Sim-to-Real Transfer}
\label{sec:sim2real}

\begin{figure}
    \centering
    \includegraphics[width=0.8\linewidth]{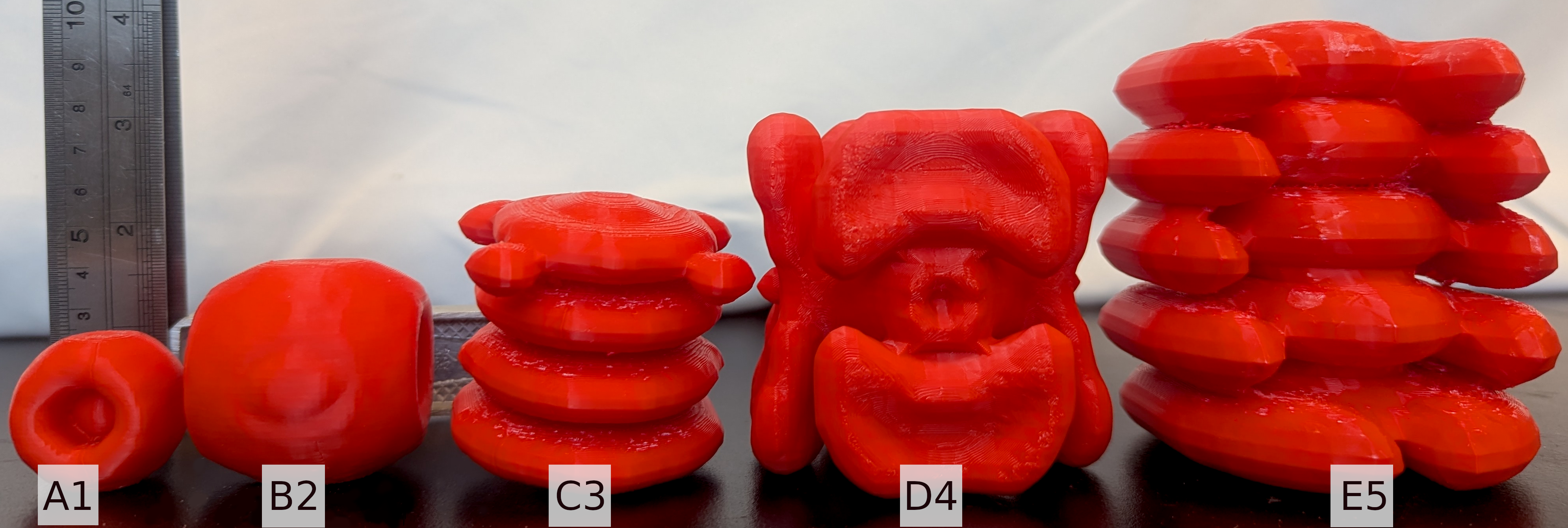}
    \caption{Labelled EGAD \cite{morrisonEGADEvolvedGrasping2020} objects used for sim-to-real translation experiments. Objects range from $\qtyrange{30}{90}{\mm}$ in bounding box edge length, $S_e$ .}
    \label{fig:egad_shapes}
\end{figure}

We deploy the $R_2$ $xyz$ policy on the hardware array without fine-tuning or adaptation. Object pose is estimated by an overhead camera array  and passed to the policy in place of the simulator state; all other observation channels are computed as in training. The policy is queried at \qty{5}{\hertz}, matching the simulated rate.

We evaluate 5 EGAD objects, \cref{fig:egad_shapes}, spanning \qty{30}{\mm} to \qty{90}{\mm}, each over the same 10 start-goal pairs drawn to cover the array, giving 50 trials. Episodes terminate on  $100$ time steps (\SI{20}{\second}) as in training. The identical (object, start, goal) triples are replayed in simulation, so the transfer can be compared, \cref{tab:sim2real}.

The $R_2$ $xyz$ policy transferred to hardware without adaptation, reaching \qty{76}{\percent} success against \qty{100}{\percent} in simulation. Degradation was seen in fine placement capabilities, as reflected in the increased median final distance. Qualitatively, failures were observed to be dominated by tips catching on concave object features. This can be seen as a direct consequence of MuJoCo resolving collisions against the convex hull; a failure mode absent from training and a true sim-to-real gap. Fine
placement depends most heavily on contact dynamics at the tip, which are also where the idealised model departs furthest from the real surface.

\begin{table}[ht]
  \centering
  \small
  \setlength{\tabcolsep}{4pt}
\caption{Sim-to-real transfer of the $R_2$ $xyz$ policy without fine-tuning: ten start--goal pairs per object on hardware, with identical pairs replayed in simulation. Trials end at $100$ steps (\SI{20}{\s}); success is a final error below \SI{20}{\mm}. \emph{Final} and \emph{Closest} give the mean planar error at the end of the trial and at nearest approach. Objects labelled by EGAD \cite{morrisonEGADEvolvedGrasping2020} identifier and $S_e$.}
  \label{tab:sim2real}
  \begin{tabular}{l rr rr rr}
    \toprule
    & \multicolumn{2}{c}{Success (\%)} & \multicolumn{2}{c}{Final (mm)} & \multicolumn{2}{c}{Closest (mm)} \\
    \cmidrule(lr){2-3} \cmidrule(lr){4-5} \cmidrule(lr){6-7}
    Object, $S_e$ & Real & Sim & Real & Sim & Real & Sim \\
    \midrule
    A1, 30mm & 80 & 100 & 18.8 & 3.5 & 14.7 & 0.8 \\
    B2, 45mm & 90 & 100 & 11.4 & 3.6 & 4.6 & 0.4 \\
    C3, 60mm & 70 & 100 & 25.3 & 3.6 & 18.3 & 0.3 \\
    D4, 75mm & 80 & 100 & 16.4 & 3.3 & 6.3 & 0.3 \\
    E5, 90mm & 60 & 100 & 24.9 & 3.3 & 15.6 & 0.6 \\
    \midrule
    \textbf{All} & 76 & 100 & 19.4 & 3.46 & 11.9 & 0.48 \\
    \bottomrule
  \end{tabular}
\end{table}

%% file: Sections/8_New_Disscussion.tex
\section{Discussion}
\label{sec:discussion}

\textbf{Scale and Geometry.} Manipulation strategy is dictated by object scale relative to the \ac{C2C}. With $S_e$ the characteristic object dimension, three regimes arise. Sub-cell objects ($S_e < d/\sqrt{3}$, the inscribed diameter of a cell) rest on the membrane alone within a single cell and are translated by deforming that cell only. Cell-scale objects ($d/\sqrt{3} \leq S_e < 1.5d$) are supported by the tips bounding one cell and move between cells by travelling deformation. Multi-cell objects ($S_e \geq 1.5d$) bridge several cells and are supported at enough points for their pose to be commanded directly, permitting gait-like transport and rotation about the surface normal - unavailable at smaller scales, where a single contact region offers no moment arm. Across these regimes the hardware manipulated objects that span \qtyrange{15}{90}{\mm}, a six-fold range, and the learned policy covered \qtyrange{30}{150}{\mm} in simulation.

Object geometry also determines which primitive is effective, most sharply seen seen between objects that roll and those that slide. The present observation space does not expose any geometric or contact information, so a general policy cannot condition on it; contact estimation, from vision or sensing at the tips, is the natural route to supplying it.

\textbf{Sampling and surface reconstruction.} The array renders a continuous surface through $N$ discrete contacts, so representable geometry is bounded by delta number and density, in analogy with multidimensional sampling \cite{petersenSamplingReconstructionWavenumberlimited1962}. The analogy is imperfect in two respects. First, the samples are mobile. Lateral delta motion makes their positions reconfigurable, so local density can be raised where finer detail is needed at the cost of density elsewhere.
Second, the membrane is the reconstruction filter, and it is neither passive nor ideal. For independent deltas the map from joint space to rendered surface would be many-to-one, owing to their lateral travel. Under tension, the membrane instead assumes the shape minimising its stored elastic energy while passing through all delta tips, so joint space is in this sense fully expressed in the rendered surface and strain. The same energy minimisation makes the surface non-local: equilibrium is reached globally, so the geometry within a cell is not determined solely by its three bounding deltas; the fabric may bow out of their plane when neighbouring deltas impose strain across the cell. Intra-cell geometry is consequently not independently commandable.
Complexity further arises when the membrane is not taut; slack regions admit wrinkling, which is multi-stable and hysteretic.

\textbf{Resolution.} The \ac{DCT} order bounds the resolution of any commanded deformation: higher orders resolve finer detail. The transform is fitted to the bounding box of the ring, so a smaller neighbourhood resolves finer detail at a given order. The minimum admissible wavelength depends on heading (\Cref{eq:lambda_min}), so a component is free of aliasing in every heading only if its wavelength exceeds $\max \lambda_{\min}^{\theta} = \sqrt{3}\,d = \qty{75}{\mm}$. 
All policies use only the first three coefficients of the \ac{DCT} so that the basis is matched across neighbourhood sizes and \cref{tab:training_runs} compares neighbourhoods at equal action dimension. However, at third order, the whole array is only able to resolve waves up to a minimum of \SI{263}{\mm}, the x axis span of the array. Sufficient to transport objects but not to shape the surface locally near the goal, consistent with its near-equal success rate and doubled median error in \cref{tab:training_runs}. Higher order coefficients may allow for finer control and a lower median error, though they come at the cost of increased dimensionality. This trade off is a promising consideration for future work.

\textbf{Departures from the ideal surface.} The nylon--spandex (80/20) fabric departs from the ideal membrane. Its loading curve is non-linear and its weave anisotropic. This along with manufacturing tolerances, compliance in the \ac{TPU} links, and motor backlash compound this deviation.  We must also consider the array boundary loads the fabric unevenly, so nominally identical configurations strain differently at the edge than at the centre. The effect of these factors combine and the result is visible deviation, such as residual ripple in a commanded plane. The associated variation in local contact geometry is absent from the idealised model and is the most likely source of the fine-placement gap observed between real and simulation in \Cref{sec:sim2real}.

The strain limit of the fabric restricts the joint workspace (\cref{fig:strain_histogram_and_curve}). For a single delta it binds only in poses with large height gradient or lateral travel, such as passing a neighbouring delta (\cref{fig:strain_workspace}); both draw on the same strain budget, so each extreme is reachable alone but not in combination. Under the biaxial strain present in the array the fabric cannot contract laterally, so the true workspace is likely smaller than modelled. Even under these limitations, the shared workspace was sufficient for every task demonstrated here. Fabrics with higher strain limits and lower anisotropy remain a promising direction for future iterations.

The ideal model also treats each attachment as a point, whereas the fabric has bonded fabric locks with finite area (\cref{fig:delta_unit}), which must be large enough for reliable adhesion. Being fixed, over each tip the fabric cannot deform, leaving patches of uncontrolled surface gradient whose extent scales with tip size. These patches matter most for objects comparable in size to the tips. A taut, unloaded membrane is harmonic and so has no interior minima; the local low points of the surface therefore lie at the tips, and small objects are funnelled towards regions from which they cannot be driven out. Domed caps prevent the tips from acting as stable equilibria, but the out-of-plane geometry they introduce can deflect objects during manipulation.

%% file: Sections/9_Conclusion.tex
\section{Conclusion}

We have presented a manipulable surface with $192$ DoF created from coupling the end-effectors of an $8\times8$ delta array with a stretchable membrane. The continuous surface removes the imposed floor of the inter-actuator spacing on object size seen by other discrete actuator arrays: objects too small to bridge adjacent tips are supported by the membrane and manipulated through its local slope, curvature, and strain, while larger objects remain driven by the tips beneath them. %

Treating actuation as a field proved productive at every level of control. Quasi-static and cyclic fields gave open-loop transport at arbitrary heading;
localised primitives routed several sub-pitch objects in parallel under closed-loop control; and a policy commanding low-order discrete cosine transform modes of a 19-delta neighbourhood halved the median error compared to commanding the whole array, transferring from simulation to hardware without adaptation. In each case, the effective command set proved smaller and more local than the actuator count suggests. Fine placement on hardware remains an outstanding gap and is the natural target for a contact-aware observation space.

%% file: delta_references.bib
@book{bohringerDistributedManipulation2000,
  title = {Distributed {{Manipulation}}},
  editor = {B{\"o}hringer, Karl F. and Choset, Howie},
  year = 2000,
  publisher = {Springer US},
  address = {Boston, MA},
  doi = {10.1007/978-1-4615-4545-3},
  urldate = {2023-12-19},
  isbn = {978-1-4613-7051-2 978-1-4615-4545-3},
  langid = {english}
}

@inproceedings{bohringerTheoryManipulationControl1994,
  title = {A Theory of Manipulation and Control for Microfabricated Actuator Arrays},
  booktitle = {Proc. {{IEEE MEMS}}},
  author = {B{\"o}hringer, K.-F. and Donald, B.R. and Mihailovich, R. and MacDonald, N.C.},
  year = 1994,
  pages = {102--107},
  address = {Oiso, Japan},
  doi = {10.1109/MEMSYS.1994.555606},
  urldate = {2024-02-20},
  isbn = {978-0-7803-1833-5},
  langid = {english}
}

@article{bradskiOpencvLibrary2000,
  title = {The {{Opencv Library}}},
  author = {Bradski, Gary},
  year = 2000,
  month = nov,
  journal = {Dr. Dobb's J},
  volume = {25},
  pages = {120--125}
}

@inproceedings{dacreScalableLowDensityDistributed2026,
  title = {Scalable {{Low-Density Distributed Manipulation Using}} an {{Interconnected Actuator Array}}},
  booktitle = {Proc. {{IEEE Int}}. {{Conf}}. {{Soft Robot}}. ({{RoboSoft}})},
  author = {Dacre, Bailey and Moreno, Rodrigo and Lambertsen, J{\o}rn and Stoy, Kasper and Fa{\'i}{\~n}a, Andr{\'e}s},
  year = 2026,
  month = apr,
  pages = {982--989},
  address = {Kanazawa, Japan},
  issn = {2769-4534},
  doi = {10.1109/RoboSoft67810.2026.11522822},
  urldate = {2026-09-02}
}

@misc{festoWaveHandling2013,
  title = {{{WaveHandling}}: {{Pneumatic}} Conveyor with Wave Motion},
  author = {{Festo AG \& Co. KG}},
  year = 2013,
  publisher = {Festo Bionic Learning Network},
  urldate = {2026-09-15}
}

@inproceedings{follmerInFORMDynamicPhysical2013,
  title = {{{inFORM}}: {{Dynamic}} Physical Affordances and Constraints through Shape and Object Actuation},
  shorttitle = {{{inFORM}}},
  booktitle = {Proc. {{ACM Symp}}. {{User Interface Softw}}. {{Technol}}. ({{UIST}})},
  author = {Follmer, Sean and Leithinger, Daniel and Olwal, Alex and Hogge, Akimitsu and Ishii, Hiroshi},
  year = 2013,
  month = oct,
  pages = {417--426},
  address = {St. Andrews, UK},
  doi = {10.1145/2501988.2502032},
  urldate = {2023-12-19},
  isbn = {978-1-4503-2268-3},
  langid = {english}
}

@inproceedings{ingleSoftManipulationSurface2025,
  title = {Soft {{Manipulation Surface With Reduced Actuator Density For Heterogeneous Object Manipulation}}},
  booktitle = {Proc. {{IEEE Int}}. {{Conf}}. {{Soft Robot}}. ({{RoboSoft}})},
  author = {Ingle, Pratik and St{\o}y, Kasper and Fai{\~n}a, Andres},
  year = 2025,
  month = apr,
  pages = {1--8},
  address = {Lausanne, Switzerland},
  doi = {10.1109/RoboSoft63089.2025.11020841},
  urldate = {2025-11-17},
  copyright = {https://doi.org/10.15223/policy-029},
  isbn = {979-8-3315-2020-5},
  langid = {english}
}

@incollection{luntzDiscretenessIssuesActuator2000,
  title = {Discreteness {{Issues}} in {{Actuator Arrays}}},
  booktitle = {Distributed {{Manipulation}}},
  author = {Luntz, Jonathan E. and Messner, William and Choset, Howie},
  editor = {B{\"o}hringer, Karl F. and Choset, Howie},
  year = 2000,
  pages = {103--126},
  publisher = {Springer US},
  address = {Boston, MA},
  doi = {10.1007/978-1-4615-4545-3_6},
  urldate = {2024-04-10},
  isbn = {978-1-4615-4545-3},
  langid = {english}
}

@article{mersereauProcessingHexagonallySampled1979,
  title = {The Processing of Hexagonally Sampled Two-Dimensional Signals},
  author = {Mersereau, R.M.},
  year = 1979,
  month = jun,
  journal = {Proceedings of the IEEE},
  volume = {67},
  number = {6},
  pages = {930--949},
  issn = {1558-2256},
  doi = {10.1109/PROC.1979.11356},
  urldate = {2026-09-07}
}

@article{morrisonEGADEvolvedGrasping2020,
  title = {{{EGAD}}! {{An Evolved Grasping Analysis Dataset}} for {{Diversity}} and {{Reproducibility}} in {{Robotic Manipulation}}},
  author = {Morrison, Douglas and Corke, Peter and Leitner, J{\"u}rgen},
  year = 2020,
  month = jul,
  journal = {IEEE Robot. Autom. Lett.},
  volume = {5},
  number = {3},
  pages = {4368--4375},
  issn = {2377-3766},
  doi = {10.1109/LRA.2020.2992195},
  urldate = {2026-09-11}
}

@inproceedings{patilLinearDeltaArrays2023,
  title = {Linear {{Delta Arrays}} for {{Compliant Dexterous Distributed Manipulation}}},
  booktitle = {Proc. {{IEEE Int}}. {{Conf}}. {{Robot}}. {{Autom}}. ({{ICRA}})},
  author = {Patil, Sarvesh and Tao, Tony and Hellebrekers, Tess and Kroemer, Oliver and Temel, F. Zeynep},
  year = 2023,
  month = may,
  pages = {10324--10330},
  address = {London, UK},
  doi = {10.1109/ICRA48891.2023.10160578},
  urldate = {2026-05-27},
  isbn = {979-8-3503-2365-8},
  langid = {english}
}

@article{petersenSamplingReconstructionWavenumberlimited1962,
  title = {Sampling and Reconstruction of Wave-Number-Limited Functions in {{N-dimensional}} Euclidean Spaces},
  author = {Petersen, Daniel P. and Middleton, David},
  year = 1962,
  month = dec,
  journal = {Information and Control},
  volume = {5},
  number = {4},
  pages = {279--323},
  issn = {00199958},
  doi = {10.1016/S0019-9958(62)90633-2},
  urldate = {2026-09-07},
  langid = {english}
}

@book{smithMathematicsDiscreteFourier2007,
  title = {Mathematics of the {{Discrete Fourier Transform}} ({{DFT}}) with {{Audio Applications}}},
  author = {Smith, Julius O.},
  year = 2007,
  edition = {Second},
  publisher = {W3K Publishing},
  isbn = {978-0-9745607-4-8}
}

@article{uriarteMethodeZurBewertung2022,
  title = {{Methode zur Bewertung der Flexibilit\"at und Wandelbarkeit am Beispiel eines omnidirektionalen F\"ordersystems}},
  author = {Uriarte, Claudio and Thamer, Hendrik},
  year = 2022,
  journal = {Logistics Journal: Proceedings},
  publisher = {[object Object]},
  doi = {10.2195/LJ_PROC_URIARTE_DE_202211_01},
  urldate = {2024-04-24},
  copyright = {fDPPL},
  langid = {ngerman}
}

@inproceedings{xueArrayBotReinforcementLearning2024,
  title = {{{ArrayBot}}: {{Reinforcement Learning}} for {{Generalizable Distributed Manipulation}} through {{Touch}}},
  shorttitle = {{{ArrayBot}}},
  booktitle = {Proc. {{IEEE Int}}. {{Conf}}. {{Robot}}. {{Autom}}. ({{ICRA}})},
  author = {Xue, Zhengrong and Zhang, Han and Cheng, Jingwen and He, Zhengmao and Ju, Yuanchen and Lin, Changyi and Zhang, Gu and Xu, Huazhe},
  year = 2024,
  month = may,
  pages = {16744--16751},
  doi = {10.1109/ICRA57147.2024.10610350},
  urldate = {2026-09-11}
}
